\documentclass[letterpaper, 10 pt, conference]{ieeeconf}  
\usepackage{graphicx}
\graphicspath{ {images/} }
\usepackage{algorithmic}
\usepackage[noadjust]{cite}

\usepackage{relsize}
\usepackage{subfigure}
\usepackage{array}
\usepackage{afterpage}
\usepackage{amsfonts}
\usepackage{mathtools}
\usepackage{amsmath}
\usepackage{hyperref}

\usepackage{amssymb}
\usepackage{relsize}
\usepackage{mathrsfs} 
\usepackage[dvipsnames]{xcolor}
\usepackage{todonotes}
\usepackage{cite}
\usepackage{color}
\usepackage{ulem} 						
\usepackage{xspace}

\usepackage{parskip}
\usepackage{color}

\IEEEoverridecommandlockouts                              

\title{\LARGE \bf
A Case Study of Spherical Parallel Manipulators Fabricated via Laminate Processes}
\author{Mohammad Sharifzadeh, Roozbeh Khodambashi and Daniel Aukes 
\thanks{The Polytechnic School, Fulton Schools of Engineering, Arizona State
University, Mesa, AZ, USA}}

\begin{document}

\maketitle
\thispagestyle{empty}
\pagestyle{empty}

\begin{abstract}
This paper evaluates how laminated techniques may be used to replicate the performance of more traditionally manufactured robotic manipulators.  In this case study, we introduce a laminated 2-DOF spherical, parallel manipulator. Taking advantage of laminating techniques in the construction of the robot can result in considerable saving in construction costs and time, but, the challenges caused by this technique have to be addressed. By using stiffer materials in rigid links, the rigidity of the robot is increased to an acceptable level.  We discuss one method for compensating position uncertainty via an experimental  identification technique which uses a neural network to create a forward kinematic model. Final results show that the proposed mechanism is able to track desired rotation with acceptable precision using open-loop model-based control. This indicates that parallel manipulators fabricated using lamination techniques can provide similar performance with prototypes made in conventional methods.
\end{abstract}

\section{INTRODUCTION}
Robotic manipulators are expensive, typically requiring the use of rigid metal links and precision components such as ball bearings at joints  to satisfy stiffness and precision requirements. Parallel mechanisms take advantage of multiple pathways to the ground, often making them more rigid and precise than their serial counterparts. Moreover, parallel robots are usually capable of achieving higher speeds due to the fact that actuators can be proximally mounted on the fixed chassis, reducing loads on distal joints. 
These benefits make parallel mechanisms a rich area for research.  Specific implementations include the Gough-Stewart platform~\cite{stewart,amir}, Delta robot~\cite{delta}, 3RRR Parallel Planar Robot~\cite{3R,3RRR_doroudchi}, 3-Degree-of-Freedom (DOF) decoupled parallel robot~\cite{Trip,mohsen} and the 2-DOF spherical parallel manipulator~\cite{kong}, which is the case study for this research. Many studies are conducted on this mechanism's workspace optimization\cite{5barlinkage,zhangworksapce}, singularities\cite{yangsingularity}, and forward and inverse kinematics \cite{kong,arian} and dynamics. This mechanism has also been the basis for camera stabilization~\cite{safaryazdi} and object tracking~\cite{hesar2014ball} applications, and two have even been used in tandem for an active vision system~\cite{agile-pair}. While none of the research on this particular mechanism uses laminated fabrication techniques for construction of the manipulator, they are a proof of general usefulness of this manipulator. 

While there are many construction methods for manipulators, few papers use laminating techniques for making high degree-of-freedom robotic manipulators. Some recent exceptions include the delta robots presented in ~\cite{millidelta,microdeltalaminated}. These papers focus on high-speed manipulators at the millimeter and centimeter scale, respectively.  Laminate fabrication techniques have already been applied to a broad number of kinematic applications, however.  Planar four-bar mechanisms have been demonstrated in flapping-wing applications\cite{Sitti2003},  5-bar spherical linkages have been previously used to drive 2DOF leg joints in micro-robotic walking applications\cite{Baisch2014}, and linear motion has been enforced using Sarrus linkages in linear actuators\cite{Goldberg2014} and assembly scaffolds\cite{Sreetharan2012}.  

This paper aims to demonstrate the possibility of fabricating manipulators using laminated techniques by tackling common challenges caused by laminate fabrication. To this end, the 2-DOF spherical parallel mechanism is designed and constructed based on laminated techniques. Next, its highly non-linear inverse kinematics is modeled by a neural network using sampled experimental data. Then, its performance is experimentally evaluated by its precision in tracking a desired rotation.

The paper is organized as follows: Section 2 describes the challenges caused by laminate fabrication.  In Section 3, a  2-DOF rotational parallel manipulator is introduced and its design and manufacturing is discussed. The inverse kinematics of the mechanism model is obtained and used for open-loop control in section 4. The paper concludes with some remarks and suggestions for future work indicated by obtained results.



\section{Design Challenges associated with laminated techniques}
\label{Sec_Challenges}
A typical laminate layup consists of different material layers which each perform separate functions.  This includes rigid material, which can be used to form rigid kinematic links, flexible material, which is used to create flexure joints at desired locations, and adhesive material -- which is used to form a monolithic laminate. One commonly-used layer ordering is (rigid, adhesive, flexible, adhesive, rigid).  The symmetric order of these materials with the rigid material on the outside has been found to reduce peeling and delamination of adhesive and flexible materials.  The rigidity of the laminate can be tuned by adjusting the thickness of the rigid layer, as well as its planar offset from the medial axis. Additionally, other material layers, such as copper for conducting electricity,  can be utilized to perform other functions, though these materials have not been used in the current work.

A large number of materials including cardboard, acrylic, fiberglass, carbon fiber -- even aluminum or steel --  can be used as a rigid layer, as long as they are compatible with the available cutting techniques -- water-jet, laser, etc -- and bond well with available adhesives. The material type and thickness of the rigid material provides a vast design space in which one can balance rigidity, weight, cost, and durability.

The adhesive layer is responsible for gluing the rigid  and flexible layers to each other, and should be selected based on adhesive compatibility with neighboring materials. The flexure layer is used in order to provide a rotational joint in laminated devices.  Thus, it must be cuttable as well as  robust against tearing and high forces, as well as exhibit a long lifetime.  Many polymers and thin metals may be used; in our study we have selected polyester as a cheap, machinable, flexible material.

There are, however, drawbacks to using laminate techniques for manipulator construction. Following are some for the  most important challenges:

First, is the finite range of flexure hinges, which is ultimately limited to $\pm$ 180 degrees, and even less when considering the thickness of the laminate. This limitation plays one of the most important roles in designing a mechanism. While using a ball bearing solves the problem in conventional mechanism designs, in a laminated design it is limited by hinge length\ref{pic_Device_Design}. Smaller hinge lengths results in a stiffer hinge but smaller range of motion. 

A second consideration is delamination, which can occur when torsional stress or compressive forces cause separation of the laminate layers. Although layers with higher resistance to peeling could help, this is an innate issue with laminate mechanisms.

Third, is hinge twist. Using a long thin flexible hinges makes it easy to twist along axes other than the intended joint axis. While reducing hinge length can stiffen a hinge to such torsional non-idealities, this decreases the joint's range of motion, as mentioned before. An alternative `Castellated hinge' design is proposed in~\cite{doshi}.  This design reduces unwanted twist by effectively reducing the length of the flexure joint, but results in higher stress as well as a range of motion less than $\pm$ 180 degrees.

Finally, the most common problem associated with low-cost materials is durability. This problem can be found in hinges and rigid materials like cardboard. While this can be addressed through higher performance materials like fiberglass or carbon fiber, hinge durability is more challenging. A thin polyester hinge can easily tear apart, especially in high-stress situations. Hinge durability can be increased by using thicker material, but this will add unwanted rotational stiffness and damping. The alternative is to use material like fabric, which trades off torsional stability for lifetime.





\section{Design \& Construction Based on Laminated techniques}
\begin{figure}[t]
\centering
\includegraphics[scale=0.62]{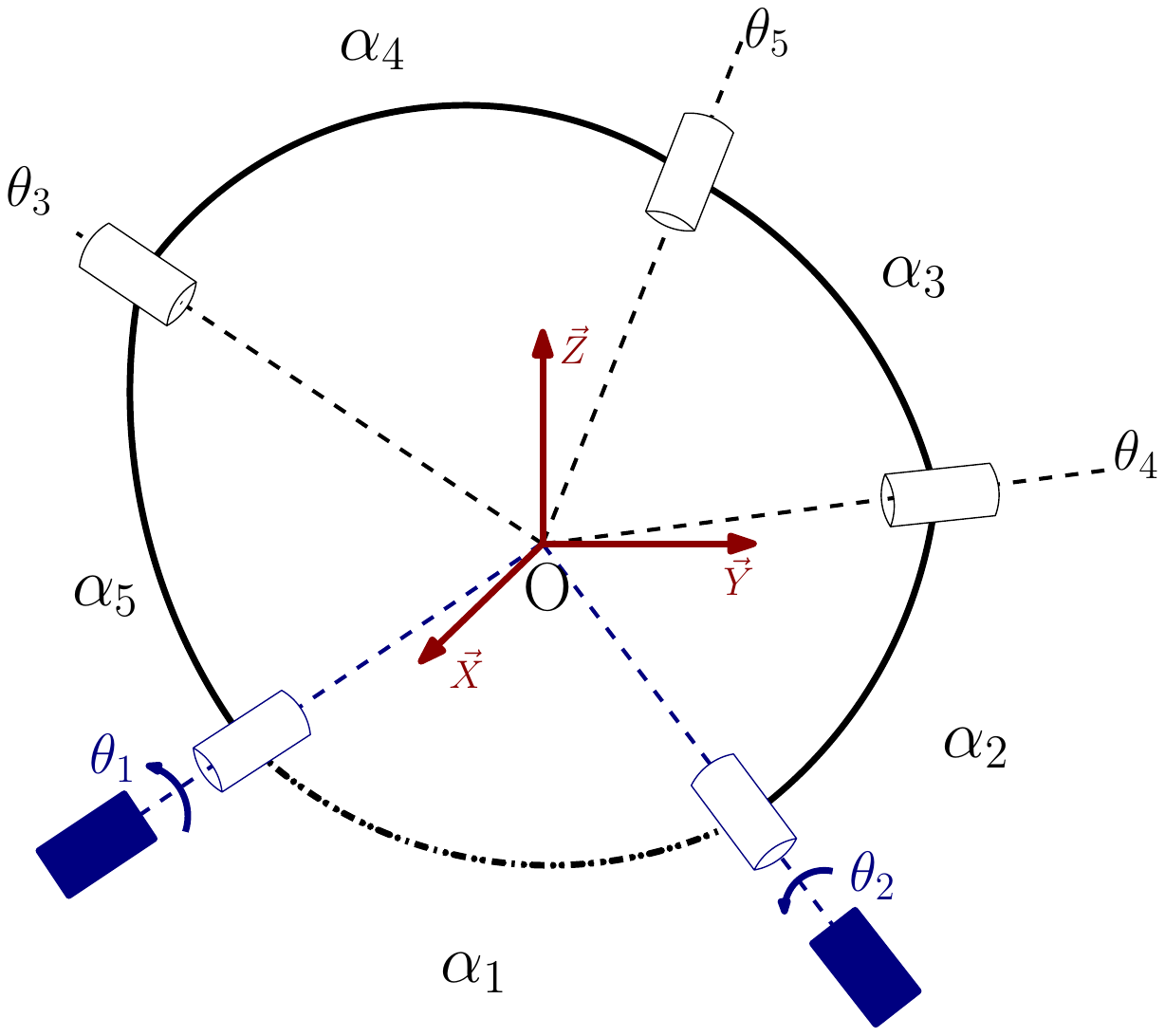}
\caption{The 2 DOF rotational linkage design. Dashed lines shows the rotation axes while the full line shows the bodies outline. The dashed dotted line represent the fixed ground.}
\label{pic_concept}
\end{figure}
\subsection{Mechanism Synthesis}
The  system under study consists of 5 rigid links with 5 hinges in between (Fig.~\ref{pic_concept}). The axes of all hinges meet at a single point (Point O) in order to form a spherical linkage. The loop closure between the two distal, rotational links requires three constraints, resulting in a two degree-of-freedom system.  By placing the end-effector at the spherical mechanism's origin, that end-effector is effectively grounded at a fixed point to the world, enabling the output motion of that frame to be represented as pure rotation.

\begin{figure*}[t]
\centering
\includegraphics[scale = 0.9]{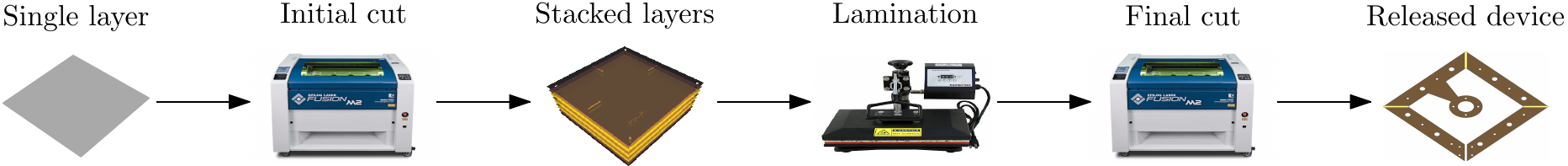}
\caption{Procedure of making a laminated device using PopupCAD as design program and heat press and laser cutter as machinery.}
\label{pic_Procedure}
\end{figure*}

The transmission which relates the actuator's motion to the end-effector's rotation may be computed using design parameters ($\alpha_i$) and the rotation between adjacent body frames ($\theta_i$), making the length scale of each link immaterial. Prior work by Ouerfelli et al demonstrates mathematically that the workspace of the mechanism is maximized if $\alpha_2=\alpha_3=\alpha_4=\alpha_5=\pi/2$  \cite{5barlinkage}. 

Based on these design parameters, the final design for the robot may be seen in Fig.~\ref{pic_Device_Design}. The mechanism's actuators are aligned along $x$ and $y$ axes of the device's chassis, considered the world frame.

Although the angle corresponding to the grounded body ($\alpha_1$) does not affect the workspace of the mechanism, it impacts the mechanism's singular states. If $\alpha_1=0$, the mechanism is singular across all inputs. Alternately, if $\alpha_1=\pi/2$ the mechanism is singularity-free~\cite{5barlinkage}. Based on that knowledge, the value for all $\alpha_1$ has also been set to $\pi/2$.  

\subsection{Jacobian \& Inverse Kinematic}
Although the mechanism has two degrees of freedom, its rotation is represented across all three axes of rotation. This is also reflected in the mechanism's inverse Jacobian matrix:
\begin{equation}
    \begin{bmatrix}
        \Dot{\theta_1}\\
        \Dot{\theta_2}\\
        0
    \end{bmatrix}
    =
    \begin{bmatrix}
        \sin{\theta_3} & 0 & -\cos{\theta_3} \\
        \dfrac{\cos{\theta_5}\sin{\theta_3}}{\sin{\theta_5}}& -\dfrac{\sin{\theta_4}}{\sin{\theta_5}} & \dfrac{\cos{\theta_5}\cos{\theta_3}}{\sin{\theta_5}}\\
        -\cos{\theta_3} & 0 & \sin{\theta_3}
    \end{bmatrix}
    \begin{bmatrix}
        \omega_X\\
        \omega_Y\\
        \omega_Z
    \end{bmatrix} 
\end{equation}
where $\theta_i$ are the hinge angles and $\omega_k$ are components of the angular velocity of the end-effector.
Based on the global axes alignment with servos, the inverse kinematics can be written as:
\begin{equation}
\theta_1 = tan^{-1} (\dfrac{N_Y N_Z}{N_X^2+N_Z^2})
\label{Eq_T1}
\end{equation}
\begin{equation}
\theta_2 = tan^{-1} (\dfrac{N_X}{N_Z})
\label{Eq_T2}
\end{equation}
where $\theta_i$ are the actuator angles and $N_k$ are the components of the unit vector perpendicular to the end-effector body.
In the case of $N_Z=0$, the values of theta are as follows:
\begin{equation*}
    \theta_1 = 0, \quad\quad \theta_2 = \pi/2
\end{equation*}

\subsection{Design \& Fabrication}
The device is constructed using a laminate fabrication process (Fig.~\ref{pic_Procedure}), similar to \cite{Whitney2011,Sreetharan2012}.  This involves cutting  individual layers of material, stacking, aligning and fusing individual layers together into a composite, and then releasing the resulting hinged laminate with a secondary cut.  This multi-material, articulated mechanism can then be erected into a three-dimensional shape, have discrete components mounted to it, and actuated using off-the shelf components.  The placement of hinges as well as the computation of cuts may be performed using PopupCAD~\cite{aukes2014algorithms}, a tool which takes the multi-layer nature of this structure into account in order to compute the cut paths for each step in the fabrication process. PopupCAD makes the design and manufacturing process for laminates easier by providing functions for defining a sheet of material, numbering layers and alignment holes for lamination, and generating support and scrap geometry that consider the geometric constraints and process limitations associated with each fabrication step.



\begin{figure}[t]
\centering
\subfigure[Design of the final device]{
\includegraphics[width=4.7cm]{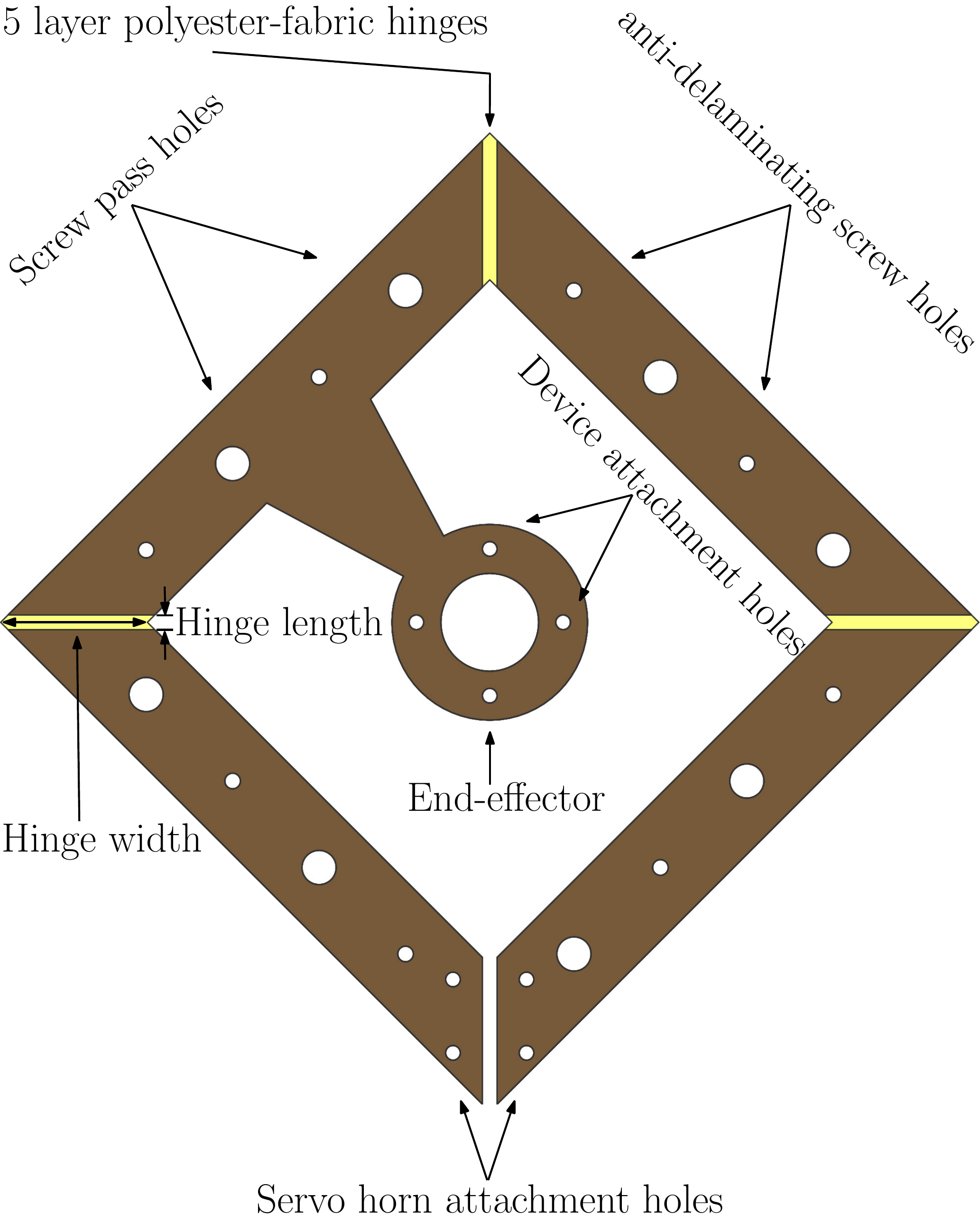}
\label{pic_Device_Design}}
\subfigure[Laminate device layer layout]{
\includegraphics[width=3.3cm]{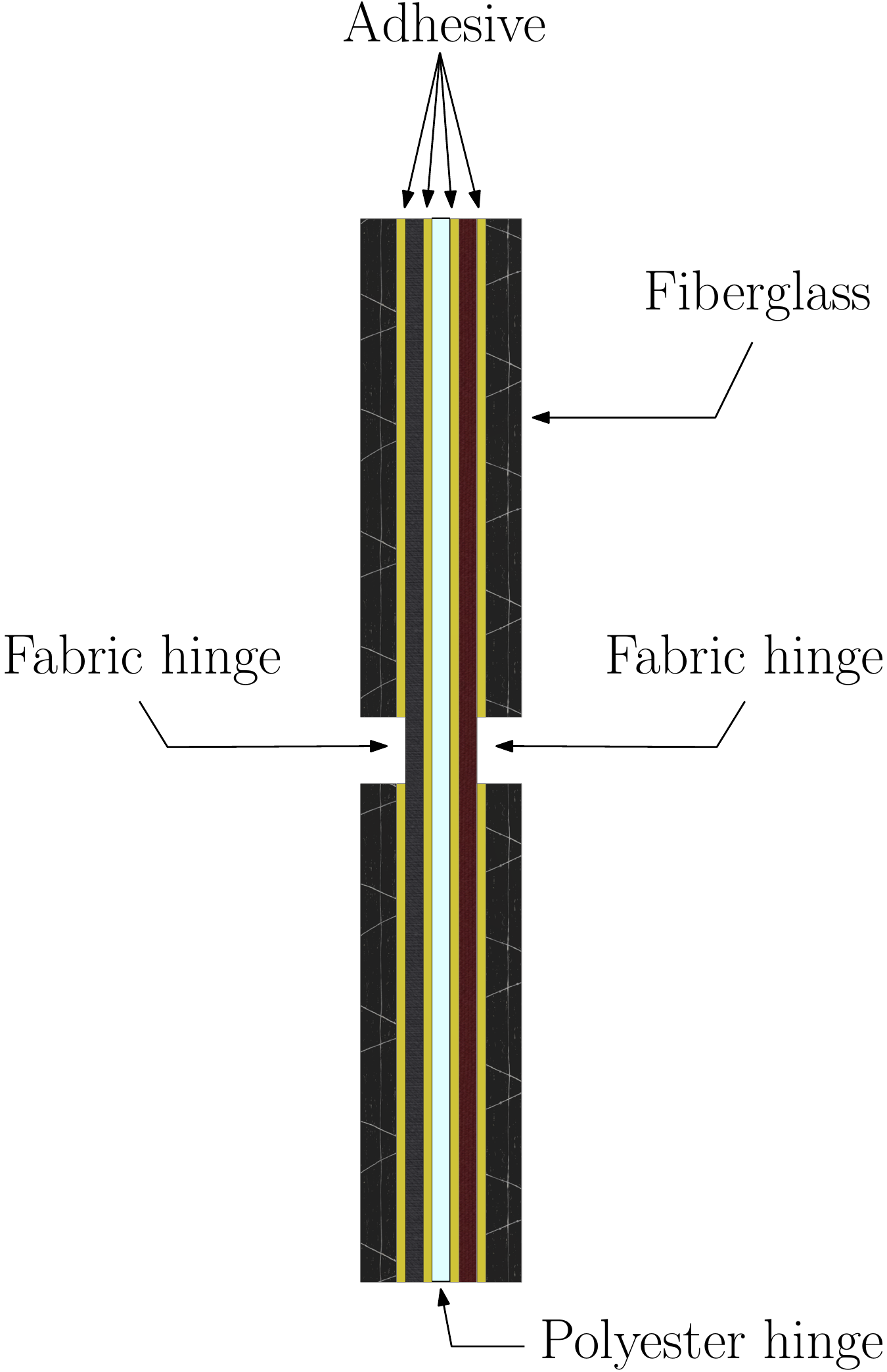}
\label{pic_LaminatedHinge}}
\caption{Design modification made in order to make the laminated design durable.}
\label{Pic_PopupCAD}
\end{figure}

Laminated fabrication techniques provide new possibilities for making articulated mechanisms. As mentioned previously, hinges can be designed with varying degrees of stiffness and damping, from a thin fabric with no stiffness or damping to a thick polyester or rubber hinges with high stiffness and damping. Elements of the hinge design such as hinge width and length directly impact the stiffness, damping, rigidity, and maximum rotation angle of the mechanism's joints (Fig.~\ref{pic_Device_Design}).  The effect of hinge design is studied in~\cite{khodambashi,doshi}.

\begin{figure}[t]
\centering
\includegraphics[scale=0.41]{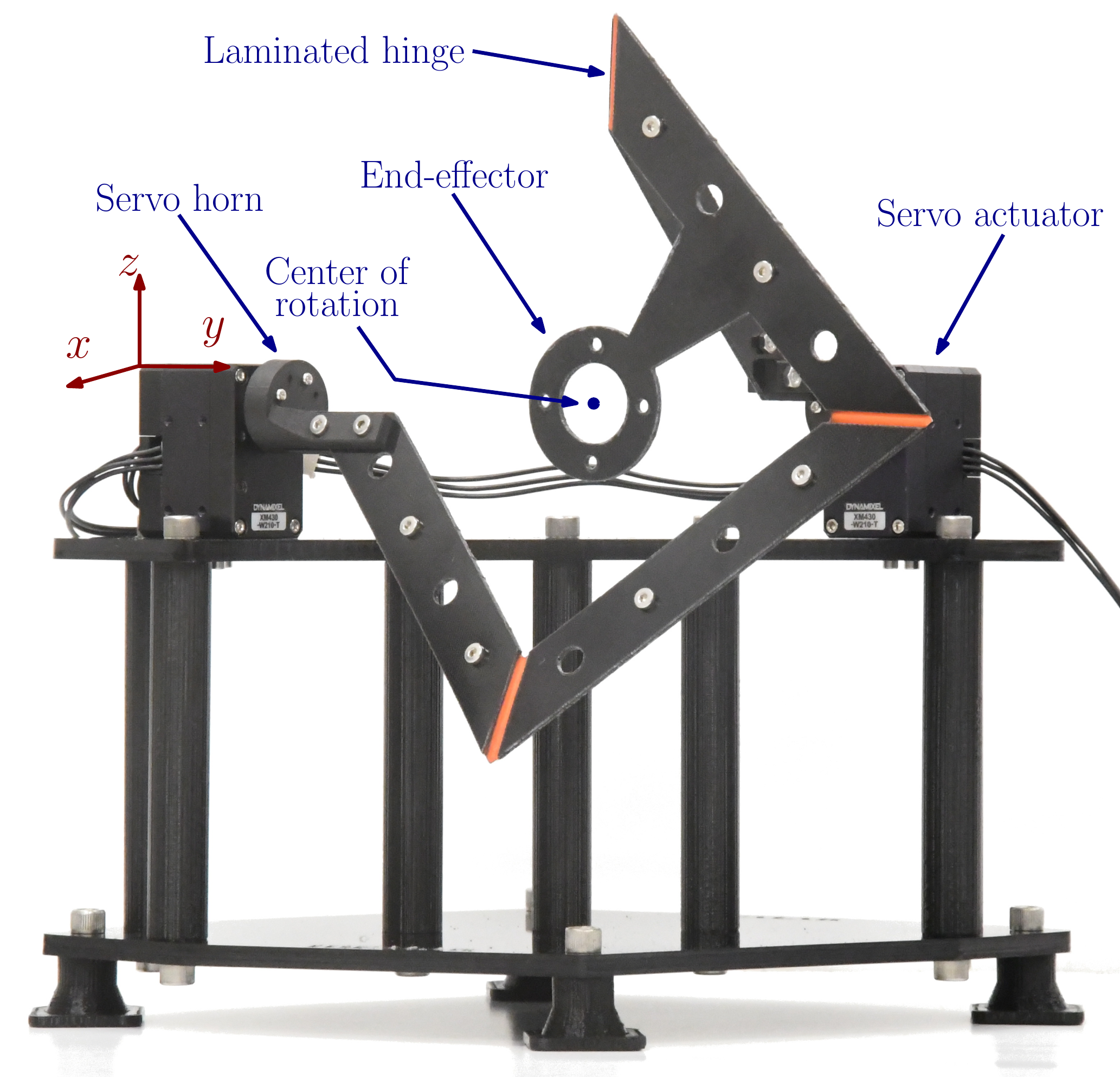}
\caption{Laminated 2-DOF rotational parallel mechanism}
\label{pic_Constructed}
\end{figure}

In this research a we propose a 5-layer polyester-fabric hinge to overcome all the challenges associated with laminated hinges mentioned in~\ref{Sec_Challenges}. By adding two outer layers of fabric to a polyester hinge, the durability of the hinge highly increases without adding unwanted stiffness and damping to the hinge. Moreover, by tuning the thickness of polyester, the long hinge twist is controlled via the material's stiffness. The length is tuned to permit the maximum range of motion ($\pm 180\deg$).

Holes are added to links in order to use rivets bolts to prevent delamination (Fig.~\ref{pic_Device_Design}) around the joints. Clearance holes are added in adjacent bodies to maintain the full range of motion (Screw pass holes in Fig.~\ref{pic_Device_Design}).

By addressing the aforementioned challenges in design (Fig.~\ref{pic_Device_Design}), the final prototype is built using 0.03-inch fiberglass sheets as rigid layer and 0.005-inch polyester sheet as polyester flexible layer. A heat-activated acrylic adhesive from Drytac\footnote{\hyperlink{http://www.drytacstore.com/mounting-adhesives/mhatm-multi-heat-adhesive.html}{www.drytacstore.com}} is used to bond layers. Figure~\ref{pic_Constructed} depicts the final prototype. Two XM430 Dynamixel DC servos are used as actuators. Two custom-made Nylon 3D-printed horns are responsible for attaching and aligning mechanism hinges to servos. The servo horns act as a safety coupling in the mechanism. The chassis is built from acrylic and 3D printed parts.




\section{Experimental Identification \& Open-loop Control}

This section explains our experimental identification and control of the 2-DOF spherical, parallel mechanism in order to evaluate its ability to achieve a precise motion profile.  The mechanism's motion has been studied using six OptiTrack cameras with a maximum frame-rate of 360 fps. The tracking system provides orientation data for specified reference frames using quaternions.  The overall process is illustrated in Fig.~\ref{pic_Ident_Process}.


\subsection{Data Gathering}
In order to capture data relating the input-output relationship between the actuators and the output frame of the end effector, the end-effector's orientation ($Q_1$) is sampled relative to the orientation of a fixed frame on the mechanism's chassis ($Q_0$) while the mechanism's actuators rotate in space. The servos are commanded with distinct sinusoidal velocity profiles rather than position commands, in order to ensure the smoothest motion profile possible.  Then, the angular position of the servos along with the quaternion data of the input and output frames is recorded and analyzed.

\begin{figure}[t]
\centering
\subfigure[Identification of mechanism inverse kinematic]{
\includegraphics[scale = 0.41]{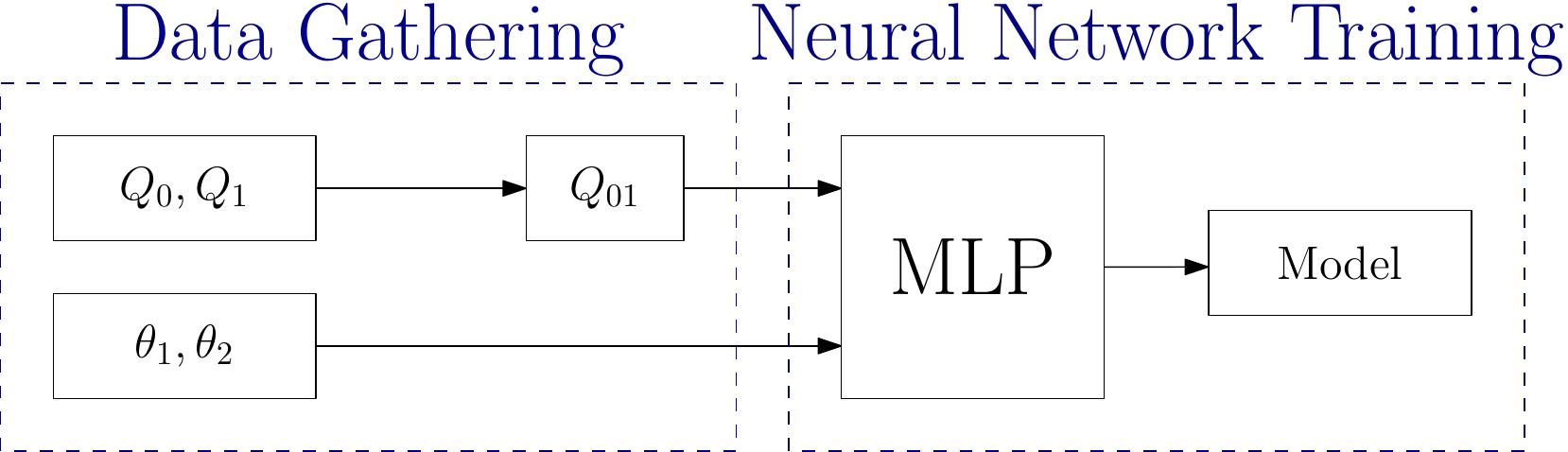}
\label{pic_Id_Process}}
\subfigure[Open-loop control of mechanism rotation]{
\includegraphics[scale = 0.41]{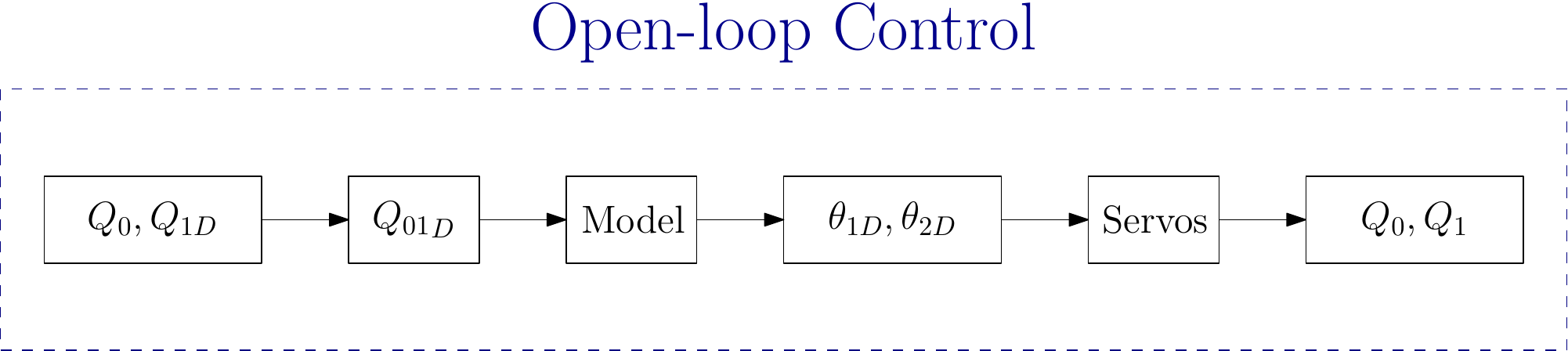}
\label{pic_Cont_Process}}\caption{Experimental identification and control process which includes data sampling, neural network training and open-loop control of the mechanism using neural network model.}
\label{pic_Ident_Process}
\end{figure}

The goal of our system identification is to model the inverse kinematics of the mechanism.  In order to compute this by processing mechanism motion data, the relative quaternion rotation ($Q_{01}$) between the fixed frame ($Q_0$) and end-effector ($Q_1$) is calculated. This relative quaternion is stored, to be used later as model input data for the identification process, while the servos' actual angles are saved as model output data (Fig.~\ref{pic_Id_Process}).  Although the theoretical solution of the inverse kinematics of the mechanism is provided in Eqs.~\ref{Eq_T1} and \ref{Eq_T2}, the goal of achieving a precise rotational mechanism requires experimental identification and modeling due to the presence of construction imperfections, non-rigid links, and flexure joints not behaving as ideal pin joints.  In order to provide a rich set of data in order to understand this complex and nonlinear set of relationships, the rotation of the end-effector and servos was sampled across 11 sets of unique velocity profiles. Using high-speed sampling, this resulted in 107,000 data points throughout all of the 11 paths. $80\%$ of the sampled data was then used to train a neural network, while the rest was used as test data.


\subsection{Identification \& Modeling}
In order to obtain a model, a multi-layer perception (MLP) neural network was trained using collected motion data. The MLP used in this study has 2700 hidden layers and uses `Adam', an optimization method proposed in \cite{adam} for solving. $\tanh$ is used as the activation function for the hidden layer. The tolerance and maximum iteration is set to 0.001 and 1000 respectively. The overall training time for the MLP process is around 5 minutes.

\begin{figure}[t]
\centering
\subfigure[Prediction of first actuator angle ($\theta_1$)]{
\includegraphics[scale = 0.37]{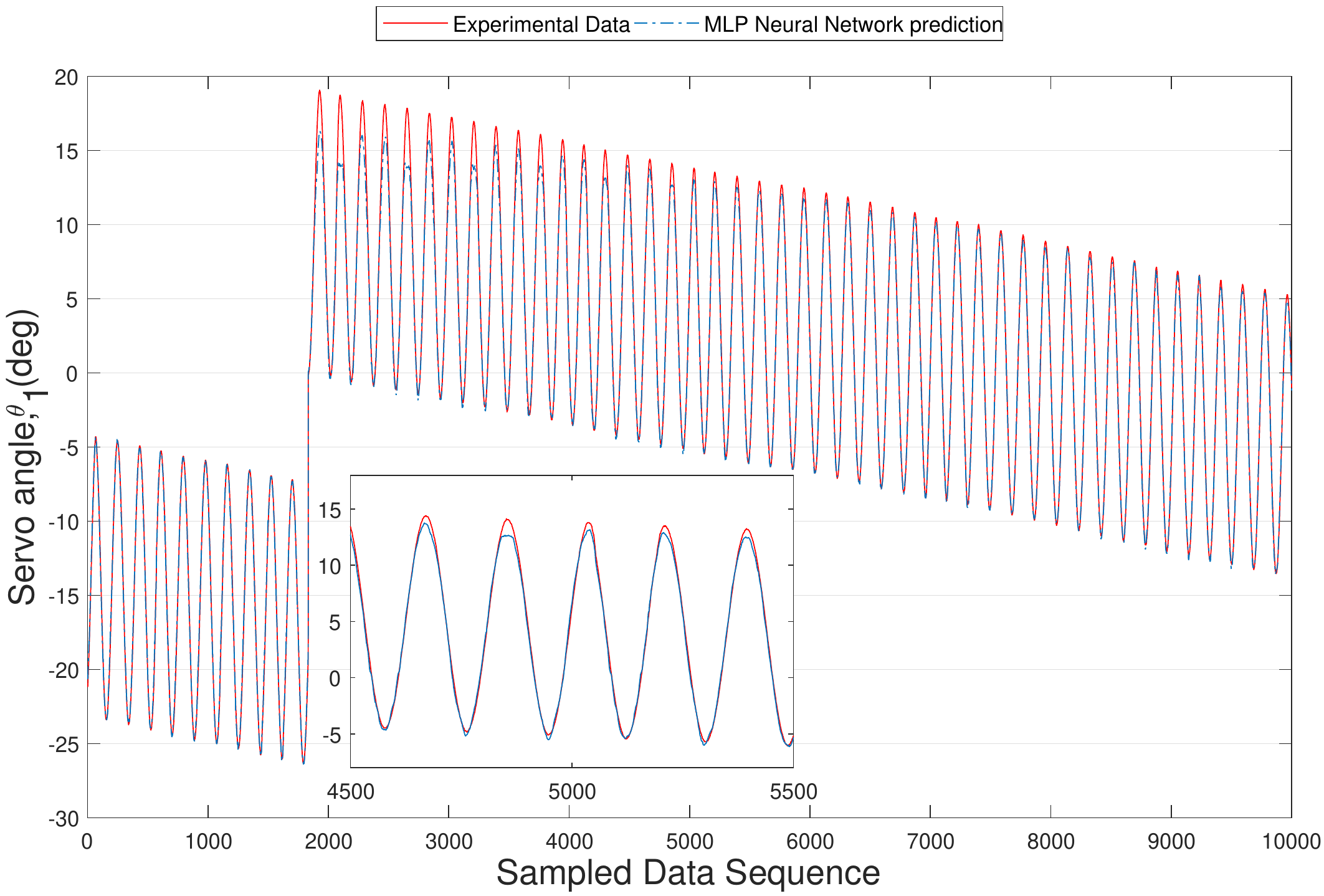}
\label{pic_NNT1}}
\subfigure[Prediction of second actuator angle ($\theta_2$)]{
\includegraphics[scale = 0.37]{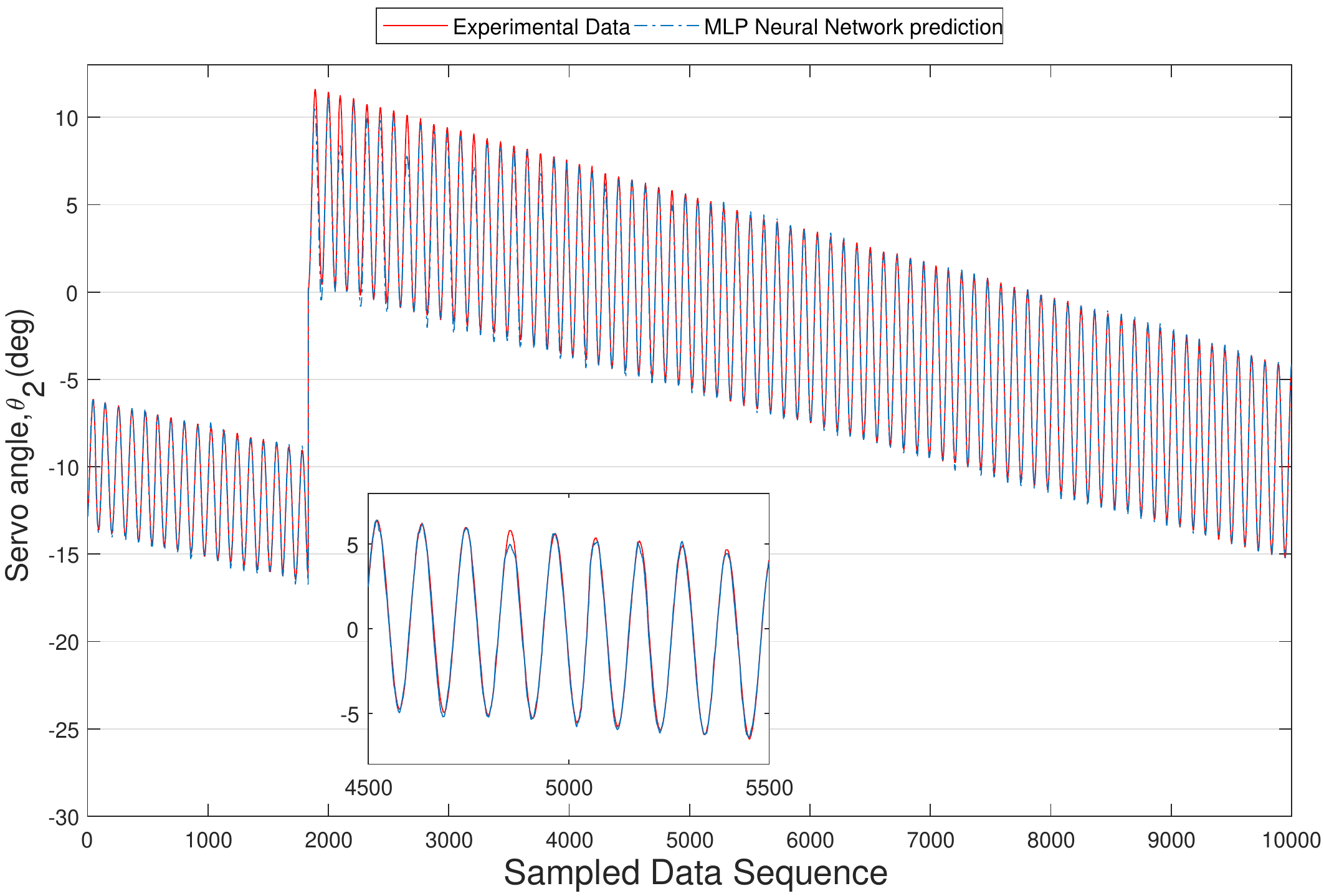}
\label{pic_NNT2}}
\caption{Performance of MLP neural network regarding test data}
\label{Pic_NN_Test}
\end{figure}

Once the MLP neural network is trained, its prediction precision is evaluated regarding the sampled test data. Figures~\ref{pic_NNT1} and \ref{pic_NNT2} show the model prediction in comparison to the real sampled data regarding fist servo angle $\theta_1$ and second servo angle $\theta_2$, respectively, for a fraction of the test data. The mean absolute error (MAE) of the model for $\theta_1$ and $\theta_2$ are $0.61\deg$ and $0.42\deg$, respectively.

\subsection{Open-loop control}
\begin{figure}[t]
\centering
\subfigure[End-effector orientation]{
\includegraphics[scale = 0.49]{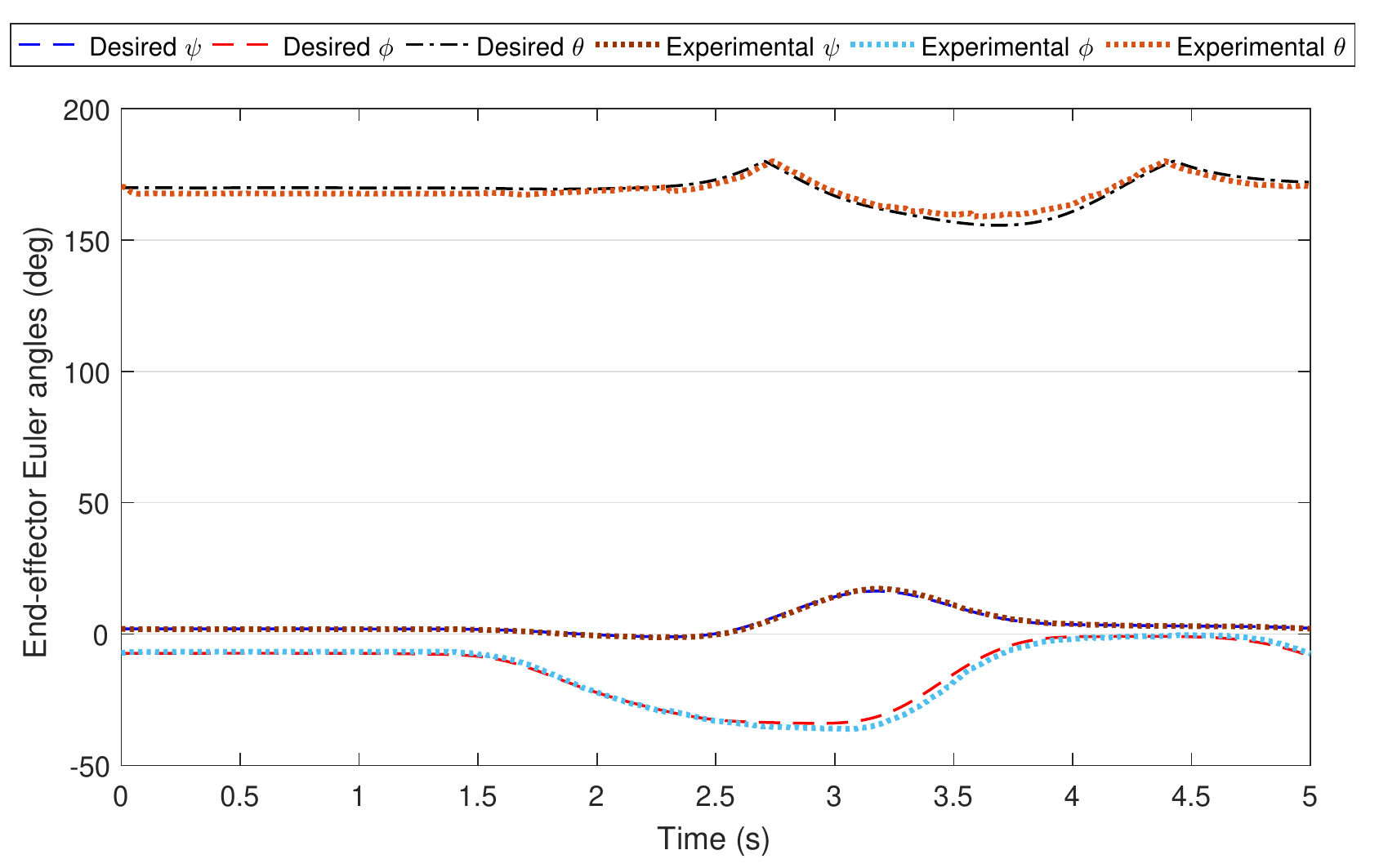}
\label{pic_Valid_Out}}
\subfigure[Position of the virtual end-effector]{
\includegraphics[scale = 0.29]{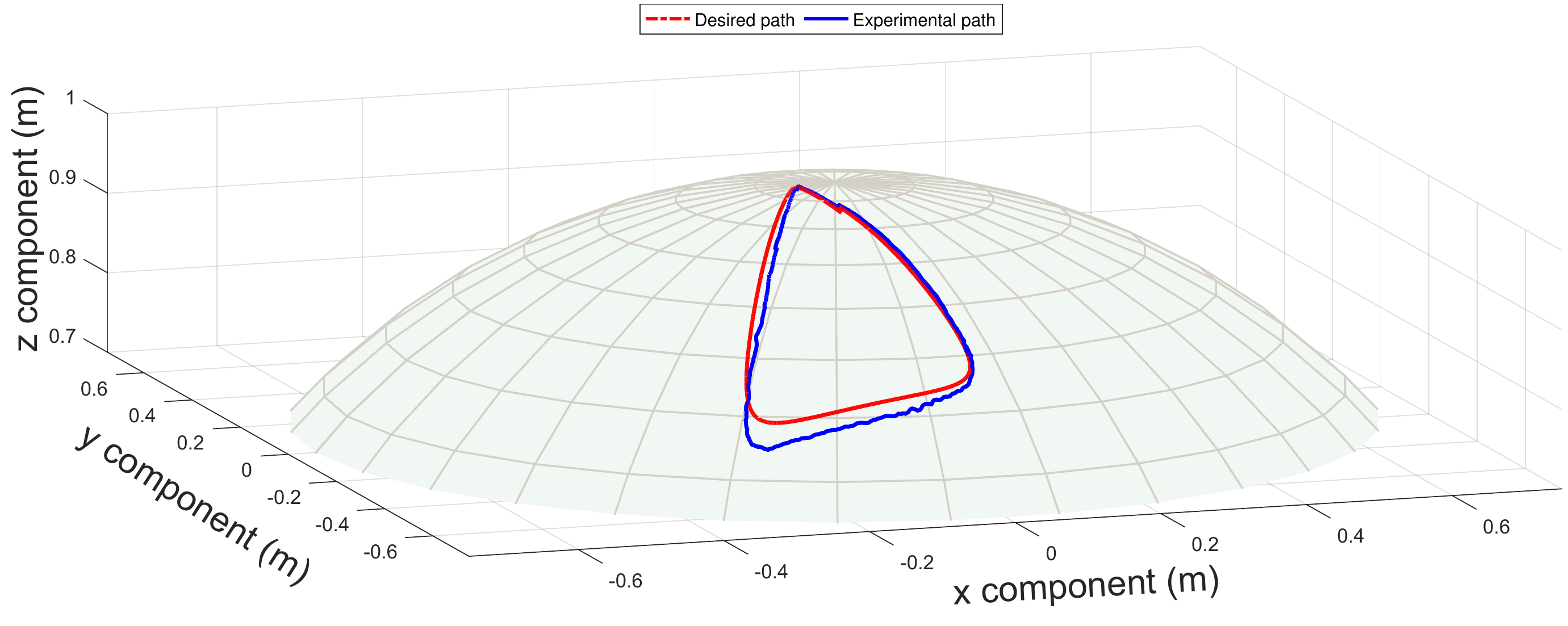}
\label{pic_Valid_Pos}}
\caption{Experimental validation of precision of the mechanism movement in presence of open-loop control based on MLP neural network.}
\label{pic_Valid}
\end{figure}
The goal of this section is to experimentally verify the precision and capability of the laminated 2-DOF rotational mechanism. 

In order to properly command the desired end-effector rotation ($Q_{1_D}$), first the desired relative orientation between end-effector and present fixed body ($Q_0$) are calculated. Then the desired relative rotation ($Q_{01_D}$) is given to the model to predict the values for desired servo angles ($\theta_{1_D}, \theta_{2_D}$). The servos are then commanded to go to desired angles. Finally, the actual orientation of the end-effector ($Q_1$) is sampled. 

While the training time of the MLP neural network is time consuming, calculation time for predicting one output is 0.5 millisecond. Hence, the frequency of the open-loop control sequence is close to 200Hz (including sending position commands to servos and sampling servo's actual position and orientation data from camera).

Figure~\ref{pic_Valid_Out} depicts the desired and actual experimental rotation of the end-effector following a desired rotation command in Euler angles. The MAE of the Euler angles of $\phi$, $\psi$ and $\theta$ are 0.24, 1.06 and 1.96 degrees, respectively. Figure~\ref{pic_Valid_Pos} displays the position of a virtual end-effector along the output frame's $z$ axis. 

Several sources of error may be attributed to the differences between goal and actual position.  This includes camera error, neural network model error, servo inner control loop position error. 




\section*{Conclusions and Future Work}

A novel, 2DOF, spherical, parallel manipulator made via laminate techniques has been introduced in this paper, based on a class of similar devices manufactured using more traditional approaches. The advantages and disadvantages of using laminate techniques have been discussed and several solutions have been proposed to address the non-ideal performance of this device, including both fabrication and modeling techniques. The paper subsequently describes the particular design investigated in this paper, including a description of the angles used and the specific fabrication choices made. Once built, the inverse kinematics of the mechanism have been derived  experimentally using a MLP neural network.  This model's ability to capture the complex and nonlinear motion of the device has been studied using a model-based open-loop control strategy to drive the end-effector through a series of desired trajectories. The resulting tracking data shows high fidelity between the predicted and experimental motion, demonstrating that our technique can not only correct for the non-ideal characteristics of a low-cost system, but to understand how to drive mechanisms without an analytical model. This technique demonstrates steps towards using low cost, durable, laminate, spherical, parallel mechanisms in place of high-precision but more expensive devices.

Laminate devices can often be manufactured faster and cheaper than conventional robots. The construction of our mechanism took less than one and a half hours and cost less than \$40.  Interestingly, reducing our mechanism's size not only makes it more rigid, but reduces overall cost.  This contrasts with conventionally-fabricated devices, where the cost increases due to the need for tighter tolerances and more precise machining. This makes laminate techniques ideal for mass production at small size scales.

In future work, we hope to use an adaptive closed-loop control algorithm in conjunction with similar learning strategies to understand the dynamics of the system in order to achieve higher-speed, smoother motion. This will enable the mechanism to adapt itself in the presence of loads on the end-effector. The next generation prototype will use metal sheets to stiffen the device further.  Future work will also study the effect of strained fabric sheets to achieve more torsion-resistant hinge designs.




\section*{ACKNOWLEDGMENT}
The authors would like to thank Dr. Mehdi Tale Masouleh and Azadeh Doroudchi for their guidance in parallel mechanism syntehesis and design, as well as Mehrdad Zaker Shahrak for his suggestions regarding neural networks.


\bibliographystyle{IEEEtran}
\bibliography{Mybib,danbib}

\end{document}